\documentclass[11pt, letterpaper]{article}

\usepackage{graphicx}
\usepackage{amsmath}
\usepackage{amssymb}
\usepackage{float}
\usepackage{setspace}
\usepackage{epstopdf}
\usepackage{color}
\usepackage[letterpaper,left=1in,right=1in,top=1in,bottom=1in]{geometry}%extra
\usepackage{multirow}
\usepackage{longtable}
\usepackage{rotating}
\usepackage{setspace}
\usepackage{geometry}
\usepackage{graphicx}
\usepackage{makeidx}
\usepackage{setspace}
\usepackage{epstopdf}
\usepackage{ragged2e}
\usepackage{caption}
\usepackage{hyperref}%extra

\usepackage{mwe}
\usepackage{mathrsfs}
\usepackage{verbatim}

\usepackage{subcaption}
\usepackage{amsmath}

\usepackage{algorithm,algcompatible,amsmath}
\usepackage{algpseudocode}

\usepackage{indentfirst}

\DeclareMathOperator*{\argmax}{\arg\!\max}% https://tex.stackexchange.com/q/83169/5764
\algnewcommand\INPUT{\item[\textbf{Input:}]}%
\algnewcommand\OUTPUT{\item[\textbf{Output:}]}%

\title{\Large \bf Optimizing the Switching Operation in Monoclonal Antibody Production: Economic MPC and Reinforcement Learning}
\author{
\centerline{\normalsize Sandra A. Obiri$^{a}$, Song Bo$^{a}$, Bernard T. Agyeman$^{a}$, Benjamin Decardi-Nelson$^{a}$, Jinfeng Liu$^{a,}$\thanks{Corresponding author: J. Liu. Tel: +1-780-492-1317. Fax: +1-780-492-2881. Email: jinfeng@ualberta.ca.}}
\vspace{5mm}\\
\centerline{\small $^{a}$Department of Chemical \& Materials Engineering, University of Alberta,} \\
\centerline{\small Edmonton, AB T6G 1H9, Canada}}
\allowdisplaybreaks

\begin{document}
\date{}
\maketitle

\begin{abstract}
Monoclonal antibodies (mAbs) have emerged as indispensable assets in medicine, and are currently at the forefront of biopharmaceutical product development \cite{rodrigues2010technological}. However, the growing market demand and the substantial doses required for mAb clinical treatments necessitate significant progress in its large-scale production \cite{farid2007process}.
Most of the processes for industrial mAb production rely on batch operations, which result in significant downtime. The shift towards a fully continuous and integrated manufacturing process holds the potential to boost product yield and quality, while eliminating the extra expenses associated with storing intermediate products.

The integrated continuous mAb production process can be divided into the upstream and downstream processes. One crucial aspect that ensures the continuity of the integrated process is the switching of the capture columns, which are typically chromatography columns operated in a fed-batch manner downstream. Due to the discrete nature of the switching operation, advanced process control algorithms such as economic MPC (EMPC) are computationally difficult to implement. This is because an  integer nonlinear program (INLP) needs to be solved online at each sampling time. This paper introduces two computationally-efficient approaches for EMPC implementation, namely, a sigmoid function approximation approach and a rectified linear unit (ReLU) approximation approach. It also explores the application of deep reinforcement learning (DRL).
These three methods are compared to the traditional switching approach which is based on a 1\% product breakthrough rule and which involves no optimization.
\end{abstract}


\section{Introduction}
Monoclonal antibodies (mAbs) are laboratory-produced antibodies designed to target specific cells such as cancerous cells in the human body \cite{nelson2000demystified}. Over the years, mAbs have gained significance as essential elements in various clinical diagnostic tests, and have been used in the treatment of diverse diseases including autoimmune diseases, cancer, and infections \cite{nelson2000demystified}, \cite{castelli2019pharmacology}, \cite{keizer2010clinical}. 

With its vital role in the pharmaceutical industry, the market size of mAbs is projected to reach an impressive \$138.6 billion by 2024 \cite{yang2019comparison}. However, the high annual cost of mAb treatments poses a significant challenge, as it can amount to approximately \$35,000 per patient \cite{farid2007process}. This cost is attributed to the required high doses and the expensive production process. Consequently, optimizing its production is imperative.

The majority of steps in mAb manufacturing rely on batch operations, such as batch/fed-batch culture and chromatography purification. 
Due to the increasing pressure to drive down cost, there is a growing interest in a transition towards a fully-continuous production process.
This transition offers potential advantages, including enhanced product quality, quantity, and stability, as well as the elimination of expenses associated with storing intermediate products. In comparison, batch operations suffer from various drawbacks such as lower efficiency, higher costs, challenges in scaling up, and compromised product quality \cite{plumb2005continuous}. 

Several studies have explored the continuous manufacturing of mAbs  \cite{yang2019comparison}, \cite{gomis2020model}, \cite{gjoka2015straightforward}. The continuous production process can be divided into two main components: the upstream process and the downstream process. In the upstream process, a perfusion bioreactor is utilized to continuously supply nutrients for cell nourishment. 
In a study by Papathanasiou et al.~\cite{papathanasiou2017advanced}, a GS-NS0 cell culture system was considered for mAb production within the bioreactor. 
To maintain a high cell density, a cell retention device is typically incorporated to recycle cells from the effluent stream of the bioreactor. The remaining contents are directed downstream for product purification through a number of chromatography steps.

To maintain continuous flow, the first step of the downstream process usually employs two or more chromatography columns. For instance, in the work by Steinebach et al.~\cite{steinebach2016model}, two columns were used to achieve the sequential countercurrent loading of mAbs for maximum resin utilization. 
Other studies have also utilized multiple columns for the same purpose, employing techniques such as simulated moving bed chromatography (SMB), sequential multi-column chromatography (SMCC), and periodic counter-current chromatography (PCC) \cite{zhang2003multiobjective}, \cite{baur2016optimal}, \cite{warikoo2012integrated}.
In this work, we focus on the use of twin chromatography capture columns, as described by Gomis et al.~\cite{gomis2020model}. The capture columns are operated such that while one column is loaded with product from upstream, the contents of the second column are  eluted into the subsequent downstream component. The columns are switched when the loading column reaches a fixed predetermined percentage product breakthrough at the column outlet. 
However, to ensure maximum product yield and effective resource utilization, the optimization and control of the switching operation is crucial.

The downstream model used in this work involves complex nonlinear equations. It also involves spatial discretization which leads to a large number of equations. Furthermore, due to the discrete nature of the switching operation, the optimization problem is formulated as a nonlinear integer problem.  This poses an extra challenge for the implementation of EMPC because an INLP needs to be solved online at each sampling time, which is computationally-demanding and slow.
To overcome these difficulties, this paper investigates and proposes computationally-efficient techniques to improve EMPC implementation. 

In the first approach, a sigmoid activation function is used to relax the discrete decision variables into continuous decision variables, making the optimization problem easier to solve. 
The second approach involves training a ReLU neural network to replace the original nonlinear model, leading to the conversion of the INLP into an integer linear program (ILP). This modification facilitates a quicker solution to the optimization problem.
Additionally, this work explores a reinforcement learning (RL) method, which seeks to identify the most effective policy for addressing the optimization problem.
The effectiveness of each method is evaluated based on product loss, cost optimization, and computational efficiency. Additionally, the impact of noise, as well as the weight factors influencing product loss and the switching action, are thoroughly investigated. The results provide valuable insights into the performance and robustness of each control approach, paving the way for enhanced downstream process control in mAb production.

%%%%%%%%%%%%%%%%%%%%%%%%%%%%%%%%%%%%%%%%%%%%%%%%%%%%%%%%%%%%%%%%%%%%%%%%%%%%%%%%%
\section{System Description}\label{Section:System description}
%%%%%%%%%%%%%%%%%%%%%%%%%%%%%%%%%%%%%%%%%%%%%%%%%%%%%%%%%%%%%%%%%%%%%%%%%%%%%%%%%%%%%%%%%%%%%%%%%%
This section provides a concise overview of the downstream production process, with a focus on the operation of the capture column.
\subsection{Overview of Production Process}
\begin{figure}[t]
	\centering
	\includegraphics[width=0.9\columnwidth]{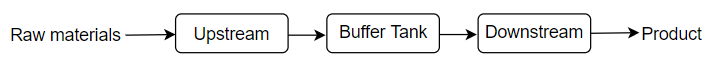}
	\caption{High-level flow diagram of mAb production.}
	\label{fig:upstream_downstream}
\end{figure}

The continuous integrated production of mAb comprises two main processes, namely, the upstream and downstream processes. As depicted in Figure \ref{fig:upstream_downstream}, a buffer tank is used between the upstream and the downstream process to ensure that disruptions in upstream operations do not significantly affect downstream operations \cite{gomis2020model}.
The upstream process is responsible for the production of mAb while the downstream process is responsible for the purification of the produced mAb. In the upstream process, nutrients are continuously introduced into a perfusion bioreactor, where a favorable environment is provided for the production of mAbs by the cells in the reactor.                         
The contents leaving the bioreactor are passed through a microfiltration unit (cell separator) to recover the cells, which are recycled back into the bioreactor. At the same time, the remaining content comprising both mAb and waste is sent downstream for further processing \cite{zhang2022smpl}. 
The mathematical model of the upstream process considered in this work is based on the studies by Kontarovdi et al. \cite{kontoravdi2010systematic}, \cite{kontoravdi2005application}.
The downstream process involves the following steps: capture, virus inactivation, cation exchange chromatography (CEX) in bind-and-elute mode and anion exchange chromatography (AEX) in flow-through mode. The downstream process configuration used in this work is taken from the studies by Gomis Fons et al. \cite{gomis2020model}.

\begin{figure}[t]
	\centering
	\includegraphics[width=0.9\columnwidth]{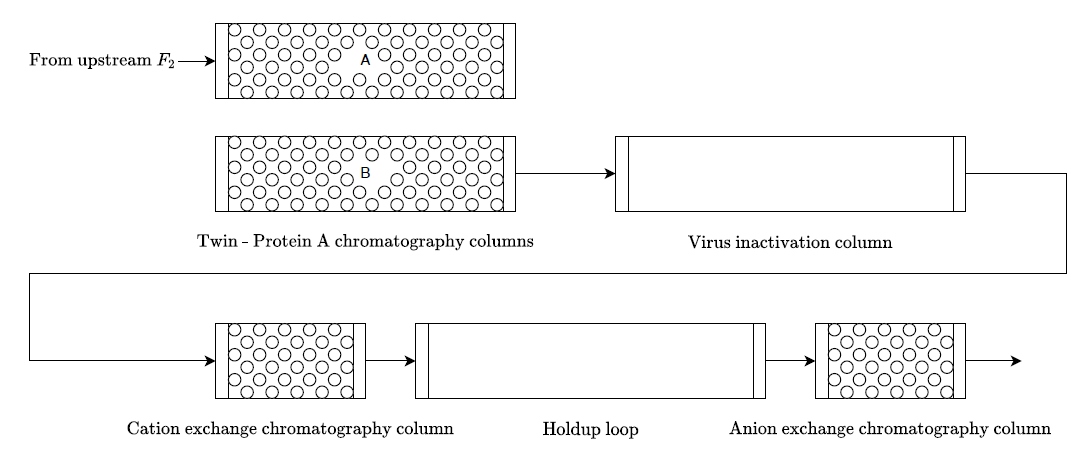}
	\caption{Schematic diagram of the downstream process~\cite{zhang2022smpl}.}
	\label{fig:downstream}
\end{figure}

The capture step is a chromatography step and is discontinuous by nature. Therefore, a set of twin protein A chromatography columns are used here to achieve continuity. This step is implemented such that while one of the twin columns (column A) is in loading mode, the other column (column B) is in elution mode as depicted in Figure \ref{fig:downstream} and the rest of the downstream processes are carried out \cite{gomis2020model}. This means that by the time column A is fully loaded, column B is empty of product and has undergone regeneration. Hence, the two columns can switch so that column A will be connected to the rest of the downstream units and its contents eluted into the VI-loop whereas column B will be connected to the upstream units and loaded \cite{zhang2022smpl}. 
Any virus that may be present is rendered inactive at the VI stage to prevent further mAb degradation. At the CEX and AEX steps, any other undesirable components or contaminants are removed from the stream before the final product is collected.

\begin{figure}[t]
	\centering
	\includegraphics[width=0.7\columnwidth]{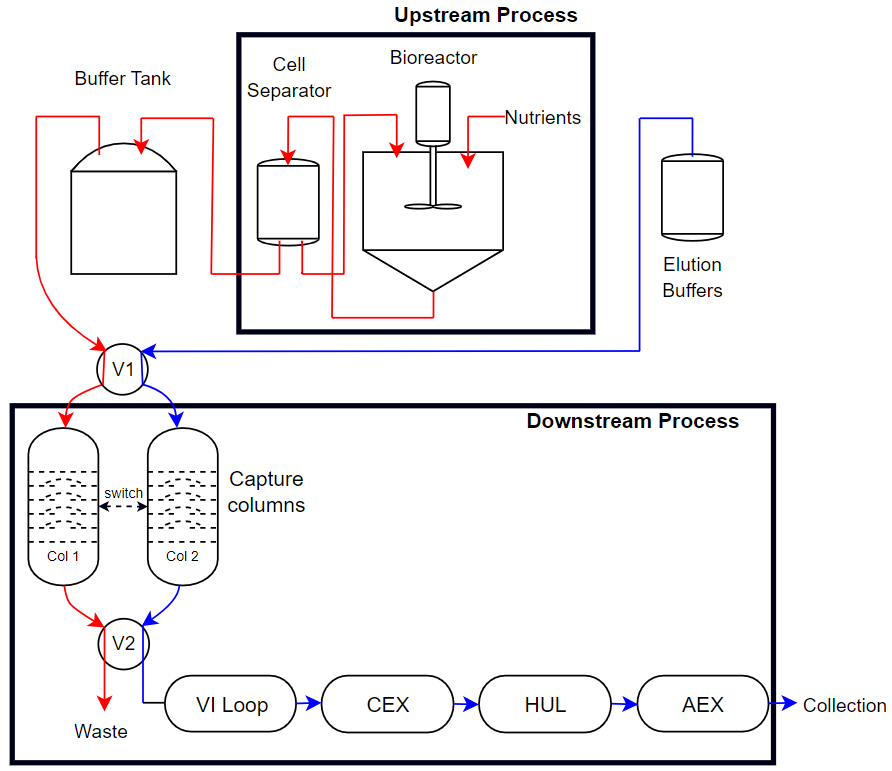}	\caption{Schematic diagram depicting the switching of capture column.}
	\label{fig:capture column}
\end{figure}

The capture step is one of the most important steps of the downstream process because the switching of the capture columns is crucial to the continuity of the integrated process. Moving forward, this paper focuses on the operation and optimization of the capture step. 
The twin chromatography columns used for the capture step are packed with protein A resins or beads, which have an affinity to the mAbs produced upstream. As the stream saturated with product flows from upstream into the capture column, the mAbs bind to the beads until the beads are saturated and no more adsorption can occur. In elution mode, a buffer is used to wash out (elute) the adsorbed mAbs from the capture column into the VI loop.
In Figure \ref{fig:capture column}, the red line represents the loading mode of the capture column and the blue line represents the elution mode of the capture column. Column 1 and column 2 are the twin chromatography columns of the capture step, and the switching of the columns is made possible with the help of the two versatile valves, $V_1$ and $V_2$.  As depicted in the diagram, any excess product from upstream passing through the capture column goes to waste when the column is not switched frequently enough. On the other hand, switching too frequently may not allow enough time for the adsorption of the mAbs onto the beads. Therefore, determining the right time to switch is important.

\subsubsection{Capture Column - Loading Mode}
\begin{figure}[t]
	\centering
	\includegraphics[width=0.6\columnwidth]{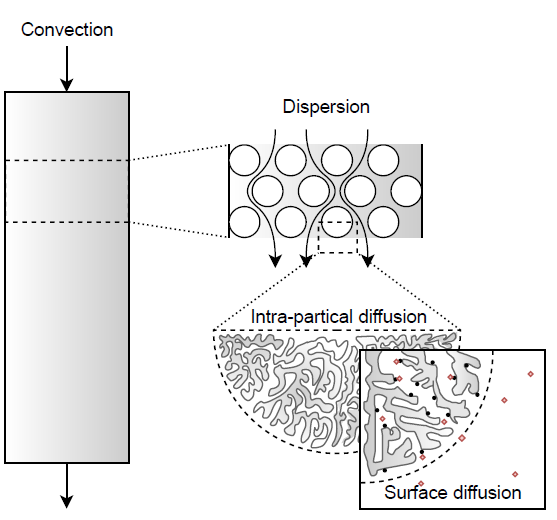}
	\caption{Schematic diagram of the chromatography column~\cite{dizaji2016minor}.}
	\label{fig:a look inside capture column}
\end{figure}

To better understand the modelling of the capture process, we must take a deeper look at the capture column as shown in Figure \ref{fig:a look inside capture column}. From the second subfigure, it can be seen that the capture column is packed with protein A resins. The porous nature of the resins provides a large surface area for the binding of the antibodies.
In the modelling of the capture step, we consider three types of mass transfers within the column. The first is the convection flow resulting from the bulk movement of the fluid through the column  in the axial direction. The second is the dispersion of mAb along the axial direction of the column as shown in the second subfigure. The third is the intra-particle diffusion within the beads \cite{zhang2022smpl}. Here, the general rate model (GRM) by Perez-Almodovar and Carta \cite{perez2009igg} is used to describe the loading of the column using the following equations:
\begin{equation}\label{eq:capture:mobilephase}
    \frac{\partial c}{\partial t} = D_{ax} \frac{\partial^2 c}{\partial z^2} - \frac{v}{\epsilon_c}\frac{\partial c}{\partial z} - \frac{1-\epsilon_c}{\epsilon_c} \frac{3}{r_p} k_f(c-c_p |_{r=r_p})
\end{equation}

This equation describes the mass transfer along the axial coordinate of the column. $z$ represents the axial coordinate of the column. $c$ represents the concentration of mAb in the mobile phase which changes with time $t$, $D_{ax}$ is the axial dispersion coefficient and $v$ is the superficial fluid velocity.  $\epsilon_c$ denotes the extra-particle column void, $k_f$ is the mass transfer coefficient and $r_p$ is the radius of the porous particles.
Furthermore, the term $\frac{\partial^2 c}{\partial z^2}$ describes the movement of mAb through the column due to the concentration difference within the column. The term $\frac{\partial c}{\partial z}$ also represents the change in concentration of mAb along the z axis due to convection flow. The last term $k_f(c-c_p |_{r=r_p})$ is used to describe the mass transfer between the mobile phase $c$ and the surface of the beads $c_p |_{r=r_p}$ \cite{zhang2022smpl}. It should be noted that the general rate model is used under the assumption that the transfer along the radial direction of the column is negligible and the transfer along the axial direction of the column and the radial direction in the beads is considered \cite{zhang2022smpl}.

The following equations are the boundary conditions for Equation~\eqref{eq:capture:mobilephase}:
\begin{subequations}
\begin{align}
    \frac{\partial c}{\partial z} &= \frac{v}{\epsilon_c D_{ax}} (c-c_F) \mbox{ ~~at~~} z=0 \label{eq:capture:mobilephase:boundary1}\\
    \frac{\partial c}{\partial z} &= 0 \mbox{ ~~at~~} z=L \label{eq:capture:mobilephase:boundary2}
\end{align}
\end{subequations}
where $c_F$ denotes the harvest mAb concentration from upstream.

The mass balance for the diffusion of mAb inside the beads is represented by the following equation:
\begin{equation}\label{eq:capture:particle}
     \frac{\partial c_p}{\partial t} = D_{eff} \frac{1}{r^2} \frac{\partial}{\partial r}(r^2 \frac{\partial c_p}{\partial r}) - \frac{1}{\epsilon_p} \frac{\partial (q_1 + q_2)}{\partial t} 
\end{equation}
where $c_p$ is the concentration of mAb along the radial coordinate of the beads, $D_{eff}$ is the effective pore diffusivity, $r$ is the distance from the current location to the center of the particle and $q_1$ and $q_2$ are the adsorbed mAb concentrations for the slow and fast-binding sites of the beads respectively.
The boundary conditions for Equation~\eqref{eq:capture:particle} are given as:
\begin{subequations}
\begin{align}
    \frac{\partial c_p}{\partial r} &= 0 \mbox{ ~~at~~} r=0 \label{eq:capture:particle:boundary1}\\
    \frac{\partial c_p}{\partial r} &= \frac{k_f}{D_{eff}} (c-cp) \mbox{ ~~at~~} r=r_p \label{eq:capture:particle:boundary2}
\end{align}
\end{subequations}

The concentration of adsorbed mAb ($q_1$ and $q_2$) are also given by the following equation:
\begin{equation}\label{eq:capture:adsobed}
   \frac{\partial q_i}{\partial t} = k_i [(q_{max,i}-q_i)c_p|_{r=r_p} - \frac{q_i}{K}] \mbox{ ~~for~~} i=1,2
\end{equation}\\
Here, $k_i$ is the adsorption kinetic constant, $q_{max}$ is the column capacity, and $K$ is the Langmuir equilibrium constant. In this work, we assume the porous beads have two binding sites; a fast-binding one and a slow-binding one. Two indices are used because we have two equations, one representing the fast-binding site and the other representing the slow-binding site \cite{zhang2022smpl}. This optimization work focuses on the loading mode of the capture column, hence the elution mode model is not considered.

\subsubsection{Compact Form of Capture Column Model}
The model equations, Equations \eqref{eq:capture:mobilephase}-\eqref{eq:capture:adsobed} are partial differential equations (PDEs). The equations were discretized to convert to ordinary differential equations (ODEs) using the two point central difference method and then the ODEs were solved using numerical methods. In loading mode, the capture column contains the distinct states, $c, c_{p}, q_{1}$ and $q_{2}$. With the beads discretised into 5 along the radial direction, the total number of $c_{p}$ states becomes 5 making the total number of states 8. The capture column was further divided into 75 equal parts along the axial direction, making a total of 600 states in the capture column. The compact form of the capture column model can be represented by Equation \eqref{eq:capture:compact}, where Equation \eqref{eq:capture:compact form2} represents the system state vector and Equation \eqref{eq:capture:compact form3} represents the continuous input vector which contains the inlet mAb concentration and flow rate which have the constant values $49.9219 mg/L$ and $21.6129 L/min$ respectively. These values were obtained from the upstream process at steady state.
\begin{subequations} \label{eq:capture:compact}
\begin{align}
    x(k+1) &= f(x(k), u^{c}(k)) \label{eq:capture:compact form1}\\
    x(k) &\in R^{600} \label{eq:capture:compact form2}\\
    u^{c}(k) &\in R^{2} \label{eq:capture:compact form3}
\end{align}
\end{subequations}
The modelling of the capture column with the switching operation is given by Equation \eqref{eq:capture:compact_switching1}
\begin{subequations} \label{eq:capture:compact_switching}
\begin{align}
    x(k+1) &= f(x(k), u^{c}(k))[1-u^{d}]\label{eq:capture:compact_switching1}\\
    u^{d}(k) &\in R^{2} \label{eq:capture:compact_switching2}
\end{align}
\end{subequations}
where Equation \eqref{eq:capture:compact_switching2} represents the discrete input vector which contains 0 and 1.

\section{Other Approaches Used for Capture Step}
The operation of the capture step as a batch process presents some limitations due to the underutilization of the stationary phase in the column while loading \cite{holzer2008multicolumn}, \cite{shukla2007downstream}, \cite{mahajan2012improving}. Due to this, the process may require the use of excessive amounts of solvent. To address these issues, as well as achieve a continuous operation, a number of continuous multi-column chromatography (MCC) configurations which usually use three or more columns for product capture have been proposed and implemented. They include sequential multi-column chromatography (SMCC) \cite{holzer2008multicolumn}, \cite{ng2014design} in which sequential loading onto three or more columns to maximize the productivity of the bioprocess is done \cite{baur2016optimal}.

Another well-established MCC method is simulated moving bed (SMB) chromatography \cite{zhang2003multiobjective} which is adopted from the concept of true moving bed chromatography (TMB). In contrast to the typical chromatography process where only the liquid phase moves through the column, in the case of TMB the solid phase also moves through the column, but in the opposite direction. Typically, the solid phase will move with a velocity bigger than that of the particles with higher affinity to the adsorbents (mAbs), but smaller than that of the particles with the lower affinity to the adsorbents (impurities). If the liquid and solid flow rates are chosen properly, it is possible to continuously recover the purified product as the feed is injected into the middle of the system. Due to the operating issues presented by the TMB method, the SMB method which simulates the counter-current movement of the fluid and solid phases through a flow switching sequence is often used instead \cite{pais1998modeling}.

Periodic counter-current (PCC) chromatography which involves the use of several columns has been used as well~\cite{warikoo2012integrated}, \cite{pollock2013optimising}, \cite{chen2021regressive}. Multiple columns may be operated in parallel, such that while one or more columns are being loaded, the others are in wash and elution modes. Nevertheless, some columns may also be idle as loading takes place \cite{badr2021integrated}. A substantial description of the PCC cycle operations can be found in the work by Warikoo et al.~\cite{warikoo2012integrated}. In \cite{warikoo2012integrated}, the PCC system used implemented up to four columns and an automated column switching strategy based on the UV absorbance difference between the feed inlet and the outlet of the four columns. 
Whereas in Gomis et. al \cite{gomis2020model} only two columns are used and the switching occurs only at 1\% of the product breakthrough curve, in \cite{warikoo2012integrated} the switching occurs at two different instances; the first occurs when there is product breakthrough and the second occurs when there is product saturation. Since four columns are used for the capture step alone, a total of five ultra-violet (UV) sensors are used to measure UV absorbance and monitor column breakthrough and saturation. The switching then occurs when the UV absorbance difference reaches a pre-determined value.

The high number of sensors and columns may result in higher capital cost and larger footprint compared to the use of two columns as is done in the work by Gomis et. al~\cite{gomis2020model}. 
For most MCC operations, the fundamental strategy employed is to load the columns beyond product breakthrough and capture the product on additional columns~\cite{gjoka2015straightforward}. Although multi-column chromatography has proven useful as a facilitative method for implementing continuous bioprocessing, it's development process may be more complicated compared to the use of only two columns for the capture step. Furthermore, it may lead to longer development or processing timelines which may conflict with the preference of biopharmaceutical companies to produce quality product in the shortest possible time~\cite{gjoka2015straightforward}. Therefore the use of twin protein A chromatography columns as was done in~\cite{gomis2020model} was considered for the capture step of the downstream process in this work.

\section{Traditional Control Method}
In many industries, profitability serves as the primary factor influencing decision-making. Therefore, it becomes crucial to adjust processes to optimize specified parameters and bring them closer to their optimal values while ensuring that environmental, safety, and customer requirements are met. In the absence of proper process control and optimization, a process may run below its maximum efficiency and may be more expensive to operate \cite{edgar2004control}, \cite{bauer2008economic}.

The conventional operation of the capture column relies on a fixed switching strategy, where the switching event takes place once a predetermined percentage product breakthrough is observed. This approach lacks optimization measures to minimize product loss and cost.
In a study conducted by Warikoo et al. \cite{warikoo2012integrated}, UV sensors were strategically placed at the capture column inlet feed and outlet. The decision to switch was determined based on the UV absorbance difference between these points, which indicated the attainment of a predefined percentage product breakthrough threshold. For instance, the study mentions a 3\% product breakthrough value \cite{warikoo2012integrated}, while another study \cite{gomis2020model} mentions a 1\% product breakthrough threshold. In this paper, a 1\% breakthrough threshold was used for the simulation of the traditional approach.

This work aims to enhance the operation of the capture step through the implementation of EMPC. However, the implementation of EMPC for integer optimization problems is known to be challenging. To overcome this challenge, we investigate two distinct approaches to improve the implementation of EMPC, namely, a sigmoid function approximation approach, and a ReLU approximation approach. Furthermore, a third approach which explores the use of reinforcement learning is also explored. The subsequent sections provide a comprehensive overview of the problem formulation for the two EMPC optimization approaches and a detailed description of all three approaches.

\section{Economic MPC Formulation}

Given that the primary control objective is to prevent unnecessary product loss, the optimization problem is formulated to minimize product loss. Switching is triggered when the mAb concentration at the capture column outlet starts to rise. The formulation of the EMPC nonlinear integer problem is presented below:
\begin{subequations}
\begin{align}
 \min_{u_{i}^{d}} \quad &   \sum_{i=k}^{k+N-1}  \left( W_{s}x_{out,i}  + W_{d} u_{i}^d \right)\label{eq:NLIP formulation}\\
    \textrm{s.t} \quad  {x}_{i+1} &= {f}(x_{i},u_i^{c})[1-u_{i}^d];i=k,..,k+N-1 \label{eq:NLIP formulation:1}\\
    {x}_k &= {x}(k)\label{eq:NLIP formulation:2}\\
   u_{i}^{d} &\in \{0,1\};i=k,..,k+N-1 \label{eq:NLIP formulation:3}
\end{align}
\end{subequations}
where Equation \eqref{eq:NLIP formulation} represents the cost function, while Equations \eqref{eq:NLIP formulation:1}-\eqref{eq:NLIP formulation:3} represent the constraints. In this context, $N$ denotes the prediction horizon, and $W_{s}$ represents the weight assigned to $x_{out}$. The variable $x_{out}$ corresponds to the 
concentration of mAb at the column outlet, which is the state of interest. This concentration is expressed as $x_{out} = C x$, where $C$ is a $1\times600$ vector of zeros with a single entry set to 1, 
indicating the outlet mAb concentration. Additionally, $x \in \mathbb{R}^{600}$ is a $600\times1$ vector as described in Equation \eqref{eq:capture:compact form2}.
$u_{i}^{d}$ represents the discrete decision variable, which is binary and has a value of 0 or 1. It represents the decision to either switch or not switch, with 0 representing the decision not to switch and 1 representing the decision to switch.
Furthermore, $W_{d}$ represents the weight on the discrete decision variable.

Equation \eqref{eq:NLIP formulation:1} represents the system state equation, which is multiplied by the term $[1-u_{i}^d]$. In this equation, $x_{i}$ denotes the system states, and $u_{i}^{c}$ represents the continuous inputs of the system.
The multiplication of the state equation by $[1-u_{i}^d]$ ensures that when $u_{i}^{d}=0$, the states continue to increase with time. Conversely, when $u_{i}^{d}=1$, the term $[1-u_{i}^d]$ becomes 0, causing the states to go to zero. This is akin to resetting or switching the column.

Practically, very frequent column switching would keep $x_{out}$ at zero, however, this is undesirable since it does not allow sufficient time for the product to reach the column outlet before switching occurs. Optimal operation requires the mAbs to have enough residence time to bind to the beads within the column during the loading process, as excessive switching would result in the underutilization of the columns. To address this issue, the penalty on the switching action, $W_{d}$, was introduced.
By adjusting the value of $W_{d}$, the frequency of switching can be controlled. Increasing or reducing $W_{d}$ and $W_{s}$ enables the fine-tuning of the switching behavior.

The EMPC optimization problem as shown in \eqref{eq:NLIP formulation}-\eqref{eq:NLIP formulation:3} is challenging to solve. In this work, we consider two approaches to approximate the EMPC optimization problem. One approach is to use a sigmoid function to approximate the integer decision variable and to convert the optimization problem to a nonlinear optimization problem without integers. Another approach we consider is to train a neural network with ReLU activation function to approximate the dynamics of the system and then convert the EMPC optimization problem to a integer linear program. 

\subsection{Sigmoid Function Approach}\label{subsec:sigmoid}
Sigmoid functions have a wide range of applications and are applied in numerous areas such as computer science, engineering, finance and physics \cite{kyurkchiev2015sigmoid}. They are bounded between 0 and 1, differentiable, and have a characteristic s-shape \cite{sharma2017activation}. They are widely used as activation functions in neural networks to convert a real number to a probability, and can be used in logistic regression to predict the outcome of binary classification problems. 

In the work by Agyeman et al.~\cite{agyeman2022lstm}, a sigmoid function was used to simplify a mixed-integer MPC problem in order to enhance the computational efficiency of an irrigation scheduler. In the work by Shao et al.~\cite{shao2017sigmoid}, a sigmoid function-based integral-derivative observer was applied to an autopilot design, and in the work by Khairunnahar et al.~\cite{khairunnahar2019classification}, a sigmoid function was used in logistic regression to classify malignant and benign tissue to improve breast cancer detection in women.

\begin{figure}[t]
	\centering
	\includegraphics[width=0.55\columnwidth]{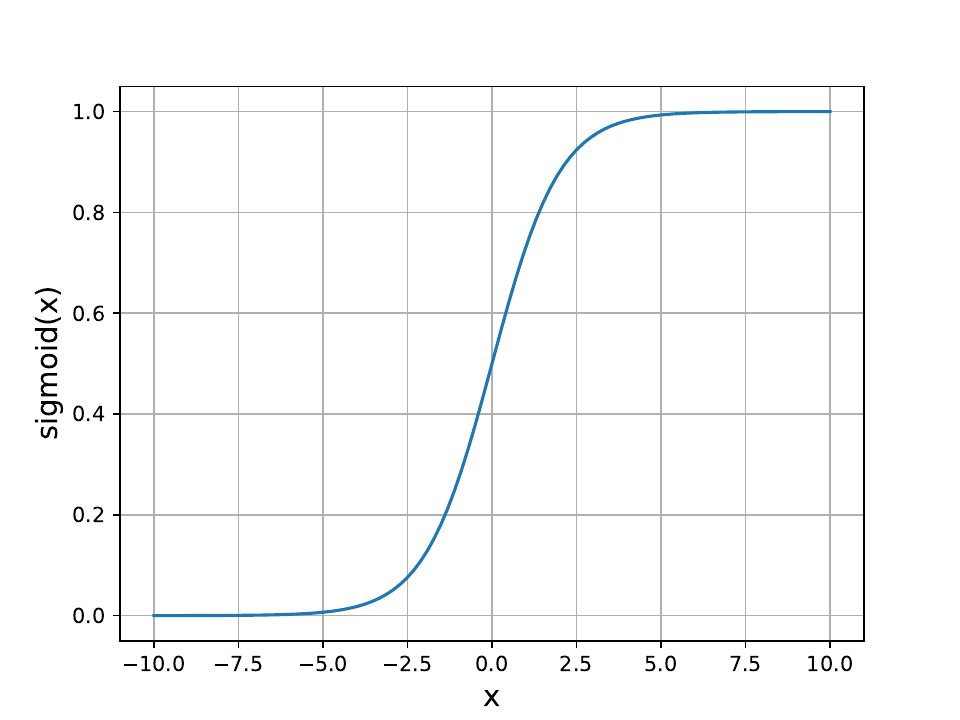}\caption{Plot of sigmoid function.}
	\label{fig:sigmoid plot}
\end{figure}

Without the implementation of the sigmoid activation function, determining the optimal discrete decision variable that minimizes the cost function would necessitate the utilization of a nonlinear integer solver like the Basic Open-source Nonlinear Mixed Integer programming (BONMIN) solver to solve the integer problem. However, this approach typically demands significant computational resources and considerable time.
Incorporating a sigmoid function allows for the utilization of continuous decision variables within defined boundaries. This transformation converts the original integer optimization problem into a continuous optimization problem, which can be addressed more efficiently using a continuous nonlinear solver such as the Interior Point Optimizer (IPOPT) solver. By solving the nonlinear optimization problem, a continuous decision variable is obtained, which is subsequently approximated as either 0 or 1 prior to its application to the actual system. The sigmoid function is mathematically defined as:
\begin{align} 
\sigma(r) = \frac{1}{1 + e^{-\beta r}}\label{eq:Sigmoid function equation2} 
\end{align} 
where $\beta$ is the slope of the sigmoid curve whose numerical value was chosen as 15, and $r$ is the continuous decision variable selected within the range $r_{min} \leq r \leq r_{max} $, where $r_{min}$ and $r_{max}$ were chosen as -10 and 10 respectively. The solver selects a continuous variable, $r$, within this range as the optimal solution. The selected $r$ value is applied to the sigmoid function, which returns an output between 0 and 1. The output is represented by $\sigma(r)$.
When $\sigma(r)$ $\geq 0.5$, it is assigned a discrete value of 1 before it is applied to the actual system. Similarly, when $\sigma(r)$ $< 0.5$, it is assigned a discrete value of 0 and applied to the system.
The optimisation problem can therefore be re-written as:
\begin{subequations}
\begin{align}
 \min_{r}  \quad &   \sum_{i=k}^{k+N-1}\left( W_{s} x_{out,i}  + W_{d} \sigma_{i}(r) \right)\label{eq:NLIP formulation3}\\
  \textrm{s.t} \quad  {x}_{i+1} &= f \left({x_{i}}, u_i^{c}\right)\left[1- \sigma_{i}(r)\right];i=k,..,k+N-1    \label{eq:NLIP formulation3:1}\\
                             r_{min} & \leq r \leq r_{max}  \label{eq:NLIP formulation3:2} \\
                               {x}_k &= {x}(k)  \label{eq:NLIP formulation3:3}
\end{align}
\end{subequations}
A plot of the sigmoid function is shown in Figure \ref{fig:sigmoid plot}.

\subsection{Rectified Linear Unit Approach}
Artificial neural networks (ANNs) are data-driven models designed to identify relationships within datasets. Unlike traditional regression models, ANNs possess the capability to learn and represent complex nonlinear relationships \cite{zou2009overview}, \cite{koray2023application}.
The architecture of an ANN consists of interconnected nodes organized into layers. The basic building block is the neuron, which receives inputs, performs a mathematical computation, and generates an output for the subsequent layer.
Neurons are activated when the weighted sum of inputs exceeds a threshold value. Once activated, the neuron applies a transfer function to the signal before passing it to neighboring nodes \cite{zou2009overview}. 

Transfer functions come in various forms, with nonlinear functions proving more valuable for handling a wide range of data patterns \cite{chiu2003genetic}, \cite{rasamoelina2020review}. While the sigmoid and hyperbolic tangent (tanh) functions have long been popular, they suffer from the vanishing gradient problem. In contrast, the ReLU transfer function has gained attention for its superior performance in training ANNs \cite{rasamoelina2020review}.

\begin{figure}[t]
	\centering
	\includegraphics[width=0.55\columnwidth]{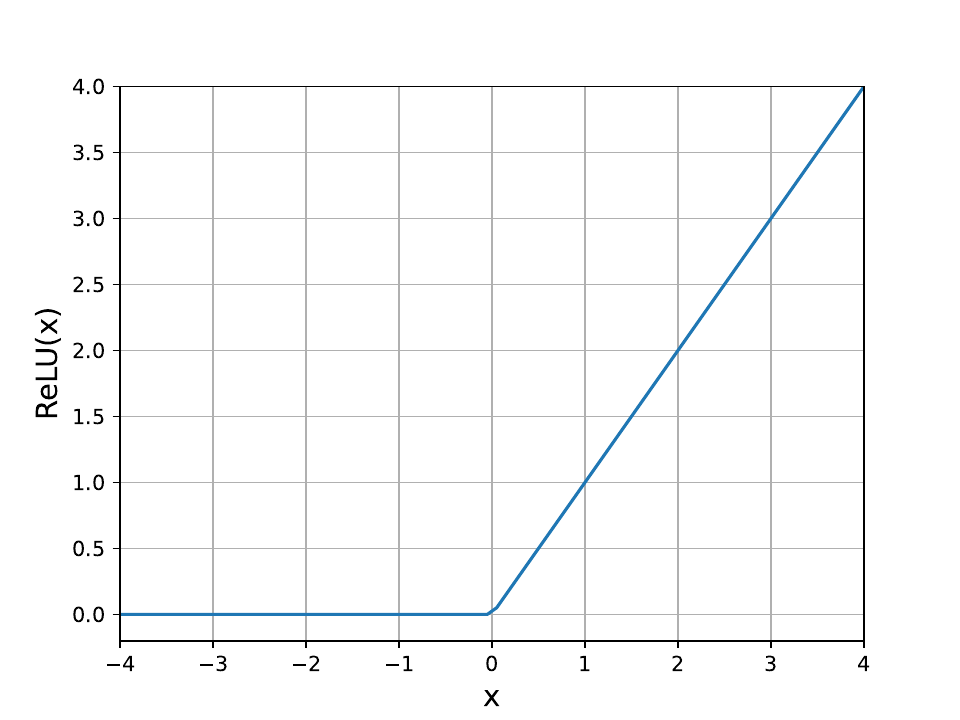}\caption{Plot of ReLU Function.}
	\label{fig:RELU}
\end{figure}

By employing the ReLU activation function, we can convert the original nonlinear optimization problem into an Integer Linear Program (ILP)  by leveraging the integer linear formulation of ReLU neural networks. This transformation, done using the optimization and machine learning toolkit (OMLT) \cite{ceccon2022omlt}, allows us to efficiently solve the resulting ILP to global optimality using readily available solvers such as Gurobi \cite{gurobi}. The ILP formulation is based on the fact that a ReLU network can be described by a set of integer linear constraints.
The ReLU activation function is defined as $ReLU(x) = \max(0,x)$, which returns the same $x$ input if $x$  is positive, and returns 0 for negative values of $x$. Figure \ref{fig:RELU} presents a plot of the ReLU function.
The ReLU neural network for this system can be represented by Equations \eqref{eq:RELU NN eqn1}-\eqref{eq:RELU NN eqn3}:

\begin{subequations}
\begin{align}
z_{0} &= 
\begin{bmatrix} 
x_{out,t}\\
x_{out,t-1}\\
u^{c_1}_{t}\\ 
u^{c_2}_{t}\\ 
u^{c_1}_{t-1}\\
u^{c_2}_{t-1}\\
\end{bmatrix}\label{eq:RELU NN eqn1}\\
z_{\ell+1} &= max\left(W_{\ell}z_{\ell} + b_{\ell}, 0\right);\ell=0,\dots, L-1 \label{eq:RELU NN eqn2}\\
y_{t}  &=  x_{out,t+1}\label{eq:RELU NN eqn3}
\end{align}
\end{subequations}
where $z_{0}$ represents the input to the neural network, which includes the state of interest $x_{out}$ and the two continuous inputs from Equation \eqref{eq:capture:compact form3} at the current and previous time steps. $z_{\ell+1}$ denotes the output vector of the $(\ell+1)$th hidden layer, and $y_{t}$ is the output of the neural network, representing $x_{out}$ at the next time step. $W_{\ell}$ and $b_{\ell}$ are the weight matrix and bias vector of the $(\ell+1)$th hidden layer, respectively.
For the $(\ell + 1)$th layer in the ReLU network described by Equation \eqref{eq:RELU NN eqn2}, let 
\underline{m}$^{\ell}$ and $\bar{m}^{\ell}$ denote the lower and upper bounds on the input such that $\underline{m}^{\ell} \leq  W_{\ell}z_{\ell}  + b_{\ell} \leq \bar{m}^{\ell}$.  The ReLU activation function can be expressed using the following integer linear constraints:\\
$z_{\ell+1} = max\left(W_{\ell}z_{\ell+1} + b_{\ell}, 0\right) \Longleftrightarrow $ 
\begin{subequations}
\begin{align*}
\begin{cases}
      z_{\ell+1} &\geq W_{\ell}z_{\ell+1} + b_{\ell}\\    
      z_{\ell+1} &\leq W_{\ell}z_{\ell+1} + b_{\ell} - diag(\underline{m}_{\ell})(1-t_{\ell})\\
      z_{\ell+1} &\geq 0\\
      z_{\ell+1} &\leq  diag(\bar{m_{\ell}})t_{\ell}\\
\end{cases}
\end{align*}
\end{subequations}
where $t_{\ell} \in$ \{0,1\} is a vector of binary variables for the $(\ell + 1)$th layer. For further information on this transformation, refer to the work by Tjeng et al. \cite{tjeng2017evaluating} and Chen et al. \cite{chen2021learning}.

\begin{figure}[t]
	\centering
	\includegraphics[width=0.6\columnwidth]{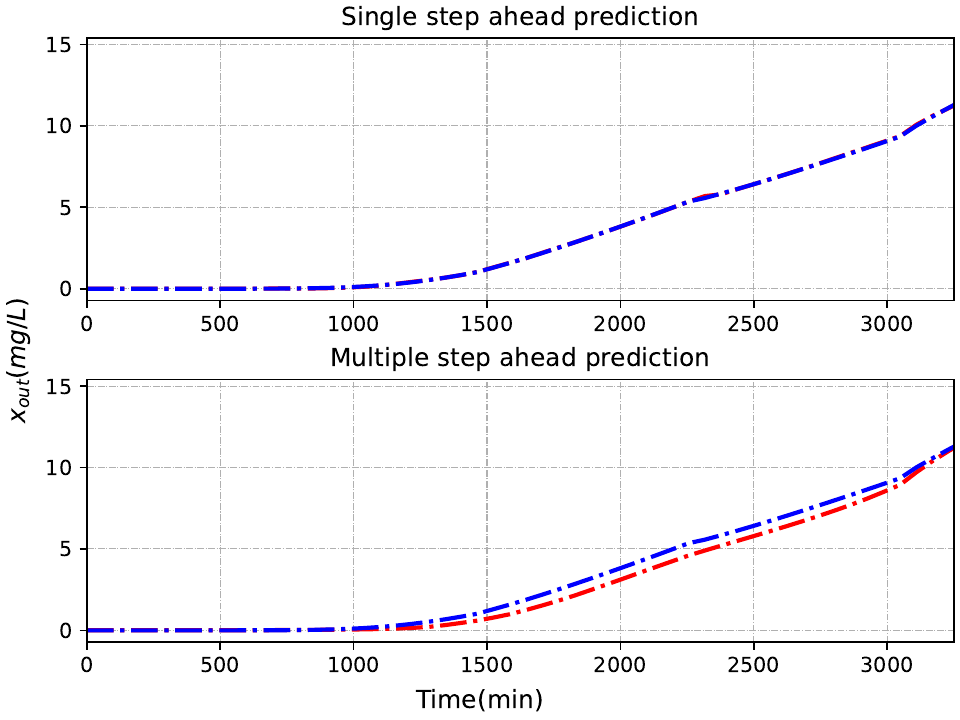} \caption{ReLU model performance.}
	\label{fig:ReLu model performance.pdf}
\end{figure}

To train the ReLU neural network, a dataset of 50,000 data points was generated for the continuous inputs $u^c(k) \in \mathbb{R}^2$ in Equation \eqref{eq:capture:compact form3}. Additionally, open-loop simulations using the first principles model described in Section \ref{Section:System description} provided a corresponding set of 50,000 data points for $x_{out}$. This input/output dataset was used to train the ReLU neural network.
The ReLU network architecture consists of two hidden layers with 200 and 160 nodes, respectively. The training process utilized current and past time steps to predict the next time step. Default learning rate and batch size values (0.001 and 32, respectively) from the Keras library were used, and a total of 80 epochs were executed. The performance of the selected ReLU model was evaluated through single-step ahead and multiple-step ahead prediction plots, as depicted in Figure \ref{fig:ReLu model performance.pdf}.
The data-driven model obtained through training the ReLU neural network is utilized by the EMPC to predict the discrete decision variable. Subsequently, the resulting solution is applied to the actual model for implementation.
The optimization problem can then be represented as:
\begin{subequations}\label{eq:Relu optimization}
\begin{align}
 \min_{{u_{i}^{d}}}  \quad &   \sum_{i=k}^{k+N-1}  \left( W_{s}y_{i+1}  + W_{d} u_{i}^d \right)\label{eq:Relu formulation}\\
    \textrm{s.t} \quad  y_{i+1} &= f_{NN}  (y_{i}, y_{i-1})  [1-u_{i}^d];i=k,..,k+N-1 \label{eq:Relu formulation:1}\\
    {y}_k &= {y}(k)\label{eq:Relu formulation:2}\\
    { u_{i}^d} &\in \{0,1\};i=k,..,k+N-1 \label{eq:Relu formulation:3}
\end{align}
\end{subequations}
where $y_{i+1}$ is the output of the neural network, and $f_{NN}$ is the neural network.

\section{Reinforcement Learning Approach}
RL represents a class of data-driven learning algorithms where an agent learns a closed-loop policy $\pi(u|x)$ by interacting with the environment. The environment is formulated as a Markov decision process (MDP), in which the agent's actions influence the system's state and result in rewards. A typical data tuple required by RL consists of the current state $x_k$, the action prescribed by the agent $u_k$, the resulting reward $r_{k+1}$, and the next state $x_{k+1}$. The state transition dynamics of the MDP are represented by the conditional probability ${P}(r_{k+1}, x_{k+1}|x_k, u_k)$. In this work, the transition dynamics are described by the nonlinear system given by Equation~\eqref{eq:capture:compact_switching}.

Following \cite{sutton2018reinforcement}, the RL problem can be formulated as finding the optimal policy $\pi^*$ that maximizes the expected return $G_k$ given the current state $x_k$ and action $u_k$:
\begin{equation}
	\pi^{*} = \argmax_{\pi} \quad {\mathbb{E}_{\pi}[G_k|x_k, u_k]}
\end{equation}
where $G_k$ represents the accumulated reward $r$. In the context of mAb switching operation, the reward $r$ is defined in Equation \eqref{eq:reward}, following the same idea of the optimization problem in Equation~\eqref{eq:NLIP formulation}.
\begin{equation}\label{eq:reward}
r_k = W_{s}x_{out,k} + W_{d}u^d_k
\end{equation}

\section{Simulation Results}
\subsection{Comparison of All Approaches}
In this section, we compare the simulation results of the four different approaches. A prediction horizon of 20 was used in the simulations involving EMPC (sigmoid and ReLU). Additionally, a step size of 60 minutes and 50 simulation steps were used across all cases, and the weights $W_s$ and $W_d$ were chosen as 1 and 0.5 respectively. For RL, the Proximal Policy Optimization (PPO) was selected as the agent. The design of the agent is summarized in Table~\ref{tb:rl design}. Due to the stochasticity nature of the RL training process, 20 RLs were trained with the same parameters, and the one with the best performance was reported in the rest of the work.
\begin{table}[t]
\normalsize
    \centering
    \begin{tabular}{|c|c|}
    \hline
    Parameters  &  Values\\
    \hline
    \hline
    Episode  &   10,000\\
    \hline
    Steps per episode     &   20\\
    \hline
    Batch size       &   10\\
    \hline
    Learning rate & 1e-4\\
    \hline
    Discounted factor & 0.99\\
    \hline
    \end{tabular}
    \caption{RL agent design parameters.}
    \label{tb:rl design}
\end{table}

% $\mathcal{}$

\begin{table}[t]
\normalsize
    \centering
    \begin{tabular}{|c|c|c|c|c|c|}
    \hline
    Method  &  Time$_{sim}$  & $\mathcal{PL}(mg/L)$  &  $\mathcal{TC}$ & Switches\\
    \hline
    \hline
    Sigmoid  &   6 h       &    0.0145  &  2.01   &  4\\
    \hline
    ReLU     &   1 h, 9 m  &    0.2421   &  1.74  &  3\\
    \hline
    RL       &   1.0713 s  &    0.3472   &   1.63 &  2\\
    \hline
    Traditional & 0.1718 s &    2.7161   &   3.71 &  2\\
    \hline
    \end{tabular}
    \caption{Comparison of all four approaches.}
    \label{tb:comparison of optimization approaches}
\end{table}
The results for all four cases are presented in Table \ref{tb:comparison of optimization approaches}, where Time$_{sim}$ represents simulation time, $\mathcal{PL}$ represents product loss, and  $\mathcal{TC}$ represents total cost.
Among the optimization approaches, the sigmoid approach had the longest simulation time, taking 6 hours to complete the closed-loop simulation of 50 steps. On the other hand, the traditional approach took only 0.1718 seconds because it involves no optimization. Furthermore, since the RL policy was obtained offline and directly implemented online, there was no optimization conducted online.  As a result, the simulation time was significantly reduced, taking only 1.07 seconds to complete.

In terms of product loss, the sigmoid approach saved more product at the column outlet compared to the ReLU and RL approaches. However, this came at the expense of higher cost, as the sigmoid approach proposed more switching actions. The ReLU approach, while saving less product than the sigmoid approach, still outperformed the RL approach in terms of product loss.  However in terms of cost, the RL approach shows the best results since it exhibited the lowest cost among the three proposed approaches.
Overall, it is evident that the proposed approaches outperformed the traditional approach since they provided better results in terms of product loss and cost optimization.

Moving forward, we decided to focus on evaluating the ReLU and RL approaches in more detail due to their significantly faster simulation times compared to the sigmoid approach. This will allow us to explore the impact of various factors on the simulation results more efficiently.
First, the effect of different EMPC prediction horizons is investigated for the ReLU approach and analyzed in Table \ref{tb:effect of horizon}. The subsequent simulation results reported in Subsection \ref{Effect of different factors-ReLU} are based on 50 simulation steps.

\subsection{Effect of Different Factors on Simulation Results - ReLU }\label{Effect of different factors-ReLU}
\begin{table}[t]
\normalsize
    \centering
    \begin{tabular}{|c|c|c|c|c|c|c|}
    \hline
     Horizon  &  Time$_{sim}$  &  $\mathcal{PL}(mg/L)$  &  $\mathcal{TC}$ & Switches &  Steps  \\
    \hline
    \hline
    1        &   2.14 m   &    1.57692  &  2.52833  &  2   &  18,36\\
    \hline
    3        &   3.76 m   &    0.55275  &  2.05275  &  3   &  16,32,48\\
    \hline
    4        &   4.54 m   &    0.55275 &  2.05275  &  3   &  16,32,48\\
    \hline
    5        &   5.11 m    &    0.24205 &  1.74205  &  3   &  15,30,45\\
    \hline
    10       &   9.60 m     &    0.24205  &  1.74205  &  3   &  15,30,45\\
    \hline
    20       &  1h, 9 m    &    0.24205  &  1.74205  &  3   &  15,30,45\\
    \hline
    30       &  2 days, 12h &    0.24205  &  1.74205  &  3   &  15,30,45\\
\hline
    \end{tabular}
    \caption{Effect of different horizons. }
    \label{tb:effect of horizon}
\end{table}

\subsubsection{Impact of Different Prediction Horizons}
The effect of different horizons on the simulation results can be analysed using Table \ref{tb:effect of horizon}. 
The results indicate that a smaller horizon leads to delayed switching, resulting in higher product loss and increased cost. On the other hand, as the horizon value increases, the controller gains the ability to look further into the future, enabling quicker decision-making. Consequently, this facilitates early switching, leading to reduced product loss and lower overall cost. For instance, for a horizon of 3, switching occurs first at the $16th$ step, whereas for a horizon of 5, the first switch occurs at the $15th$ step.

After analyzing the results, a horizon value of 10 was selected for subsequent simulations. This value strikes a balance between minimizing product loss and cost while still allowing the controller to make timely decisions based on a reasonable forecasting window.
By optimizing the horizon value, the ReLU approach can achieve more efficient control and better overall performance in terms of product loss and cost optimization.

\subsubsection{Impact of Different Weights}
\begin{table}[t]
\normalsize
    \centering
    \begin{tabular}{|c|c|c|c|c|c|c|}
    \hline
     $W_{s}$  &  $W_d$ &  Time$_{sim}$  &  $\mathcal{PL}(mg/L)$ & Switches &  Steps  \\
    \hline
    \hline
    0.9        &   0.1  &   11.27 m  &  0.09631   &  3   &  14,28,42\\
    \hline
    0.7        &   0.3   &  9.68 m  &  0.24205   &  3   &  15,30,45\\
    \hline
    0.5        &   0.5   &  9.60 m  &  0.55275   &  3   &  16,32,48\\
    \hline
    0.3        &   0.7   &  11.11 m  &  1.15787 &  2   &  17,34\\
    \hline
    0.1        &   0.9   &  11.21 m  &  1.15787   &  2   & 17,34\\
    \hline
    \end{tabular}
    \caption{Effect of weights $W_{d}$ and $W_{s}$. }
    \label{tb:Effect of weights}
\end{table}

The impact of varying the weights, $W_s$ and $W_d$, was investigated to assess their influence on the simulation results. A prediction horizon of 10 was used for the simulations, and the results were presented in Table \ref{tb:Effect of weights}, displaying the variations in simulation time, product loss, number of switches, and switching steps for different weight combinations.

A smaller switching penalty, represented by a smaller value of $W_d$, allows for more frequent switching. Consequently, when the penalty is smaller, switching commences at an earlier step, resulting in reduced product loss. As the weights differ in each case, a direct comparison of the cost is not feasible. Instead, we can focus on comparing the effects of the weights on switching and product loss. Generally, it is observed from Table \ref{tb:Effect of weights} that a higher value of $W_s$ and a lower value of $W_d$  result in a smaller number of switches and minimal product loss. In contrast, a lower value of $W_s$ and a higher value of $W_d$  lead to a reduction in the number of switches, but with an associated increase in product loss.

Moving forward, the weights $W_s$ and $W_d$ were selected as 0.7 and 0.3, respectively, based on a careful consideration of their impact on the optimization process. A weight of 0.7 was assigned to $W_s$, which controls the penalty for product loss. This weight was chosen to be relatively high, indicating the importance of minimizing product loss in the system. By placing a greater emphasis on reducing product loss, the optimization algorithm is incentivized to make decisions that prioritize maintaining product quality.

Furthermore, the weight on $W_d$ was set to 0.3, indicating the penalty associated with switching actions. This weight was selected to strike a balance between minimizing unnecessary switches and optimizing the utilization of system resources. With a moderate penalty for switching, the algorithm is encouraged to find an efficient trade-off between the number of switches and overall system performance.

By assigning weights of 0.7 and 0.3 to $W_s$ and $W_d$ respectively, the chosen values reflect a strategic approach to address the dual objectives of minimizing product loss while maximizing resource efficiency. It is expected that the selected weights will facilitate effective decision-making, leading to improved performance and cost-effectiveness in subsequent simulations and practical implementations.
Plots of the cost, $u_i^d$ and $x_{out}$ for the selected weights can be found in Figure \ref{fig:Plots of $x_{out}$, cost and $u_i^d$}.
\begin{figure}[t]
	\centering
	\includegraphics[width=0.6\columnwidth]{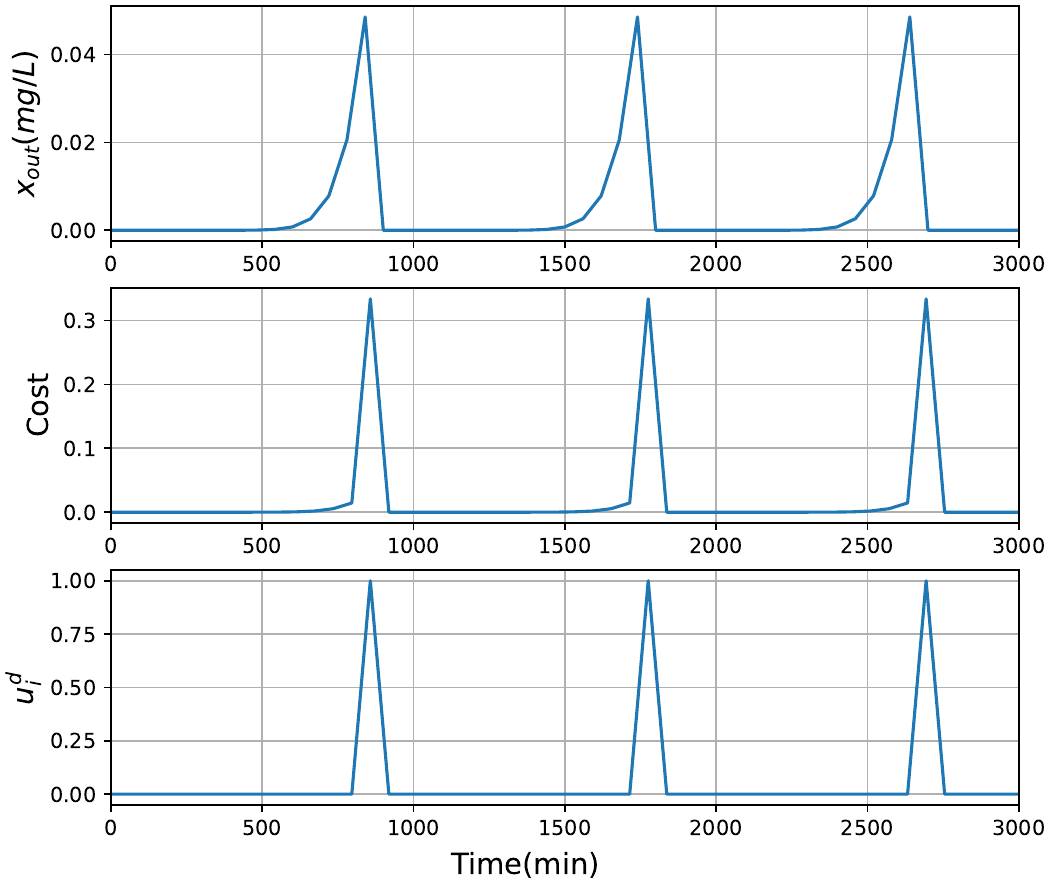} \caption{Plots of $x_{out}$, cost and $u_i^d$ for selected weights.}
	\label{fig:Plots of $x_{out}$, cost and $u_i^d$}
\end{figure}

\subsubsection{Impact of Process Noise}
\begin{table}[t]
\normalsize
    \centering
    \begin{tabular}{|c|c|c|c|c|c|}
    \hline
     Noise$_{std} $   &  $\mathcal{PL}_{ave}(mg/L)$  &  $\mathcal{TC}_{ave}$ &  Steps  \\
    \hline
    \hline
    2        & 0.19261 &   1.03482 &   14,28,42\\
             &         &           &   15,29,44\\
    \hline
    3        & 0.22476 &  1.05733  &   16,31,46\\
             &         &           &    14,29,44\\
             &         &           &    15,29,44\\
    \hline
    4       & 0.22320  & 1.05624     &   16,31,47\\
            &          &             &   13,28,43\\
            &          &             &   15,30,45\\
            &          &             &   15,29,44\\
    \hline
    5       & 0.21808  &  1.05264   &   16,32,47\\
            &          &            &   15,30,45\\
            &          &            &   15,29,44\\
            &          &            &   14,27,42\\
    \hline
    \end{tabular}
    \caption{Effect of process noise. }
    \label{tb:Impact of process noise}
\end{table}
In this section, the impact of process noise on the system performance is investigated by adding randomly-generated noise of 0 mean and different standard deviations to the two constant continuous inputs, $u_i^{c}$, defined in Equation \eqref{eq:capture:compact form3}. 
The simulations are conducted using a horizon of 10, and weights of 0.7 and 0.3 for $W_s$ and $W_d$ respectively.
The results of the simulations under the different levels of noise are summarized in Table \ref{tb:Impact of process noise}, where Noise$_{std}$ represents the standard deviation of the noise used, $\mathcal{PL}_{ave}$ denotes the average product lost, and $\mathcal{TC}_{ave}$ represents the average total cost.
For each standard deviation of noise, five simulations were run, and the average product lost and average total cost were calculated and recorded in the table.

In the presence of process noise, the switching behavior is not as perfectly periodic as it is in the noise-free cases reported in Tables \ref{tb:effect of horizon} - \ref{tb:Effect of weights}, and switching may occur a step earlier or later than anticipated. The noise introduces some level of variability in the system, resulting in fluctuations in the total product lost and total cost.
The analysis of the impact of process noise provides valuable insights into the robustness and performance of the system under realistic operating conditions. It highlights the need for appropriate control strategies to mitigate the effects of noise and maintain desired performance levels.

\subsubsection{Impact of Measurement Noise}
\begin{table}[t]
\normalsize
    \centering
    \begin{tabular}{|c|c|c|c|c|c|}
    \hline
    $\alpha$    & Noise$_{std} $   &  $\mathcal{PL}_{ave}(mg/L)$  &  $\mathcal{TC}_{ave}$  &  Steps  \\
    \hline
    \hline
    $10^{-4}$ &   2        & 0.242053 & 1.742053 &   15,30,45\\
    \hline
    $10^{-4}$ &   4        & 0.242053 & 1.203960 &   15,30,45\\
    \hline
    $10^{-4}$ &   5       & 0.242053 &  1.069437   &   15,30,45\\
    \hline
    $10^{-4}$ &   6       & 0.242053 &  1.069437   &   15,30,45\\
    \hline
    $10^{-4}$ &   10      & 0.242053 &  1.069437   &   15,30,45\\
    \hline
    $10^{-3}$ &   10       & 0.110225 & 0.977039   &  14,28,42\\
              &            &          &            &  14,28,40\\
              &            &          &            &  15,29,43\\
    \hline
    $10^{-3}$   & 12      & 0.096796 & 1.26775    & 6,16,30,40,50\\
                &         &          &            & 15,18,32,46\\
                &         &          &            & 14,17,32,43\\
                &         &          &            & 14,28,42\\
\hline
    \end{tabular}
    \caption{Effect of measurement noise. }
    \label{tb:Impact of measurement noise}
\end{table}
To analyse the effect of measurement noise, randomly generated noise of zero mean, normal distribution and varying standard deviations was added to the outlet mAb concentration, $x_{out}$. The simulation results under different levels of measurement noise are summarized in Table \ref{tb:Impact of measurement noise}, where $\alpha$ represents the multiplication factor used to adjust the intensity of the noise level.

As the standard deviation of the measurement noise increases, the system exhibits robustness in maintaining a periodic switching pattern, similar to the noise-free case. The average product lost remains relatively constant, indicating the controller's ability to adapt to the measurement noise, nevertheless, slight variations in the total cost are observed due to the influence of the noise.

As the noise level is further increased by adjusting $\alpha$, however, it can be observed that the switching pattern becomes less periodic. This suggests that the controller's performance may be impacted by higher levels of measurement noise, leading to deviations from the expected behavior.

%%%%%%%%%%%%-----------------------RL------------------%%%%%%%%%%%%%

\subsection{Effect of Different Factors on Simulation Results - RL }
In this section, the same study was conducted to examine the effects of different weights, process noises, and measurement noises on the RL controller. Because the RL design did not involve the prediction horizon, we decided not to conduct a study on it.

\subsubsection{Impact of Different Weights}
Using the same experimental design applied to ReLU with weights $W_s$ and $W_d$, we investigated the effect of five different weight pairs. The results are summarized in Table \ref{tb:Effect of weightss_rl}. Similar to the ReLU approach, an increase in the value of $W_d$ resulted in a greater penalty on the switching action, leading to a decreasing trend in the number of switches. For the last three cases, the number of switches was the same. However, as $W_d$ increased from 0.5 to 0.9, the first switch happened at a later time step, from the $18^{th}$ to the $22^{nd}$. Consequently, there were more production losses as the mAb escaped from the capture column outlet due to fewer and delayed switches. Also, we can observe from the last column that the switching was periodic because there was no disturbance considered. Moving forward, the experiments are conducted with $W_s$ and $W_d$ set to 0.7 and 0.3, respectively.
 
\begin{table}[t]
\normalsize
    \centering
    \begin{tabular}{|c|c|c|c|c|c|c|}
    \hline
     $W_{s}$  &  $W_d$ &  Time$_{sim}$  & $\mathcal{PL}(mg/L)$  & Switches &  Steps  \\
    \hline
    \hline
    0.9        &   0.1  &   0.87 s  &  0.096 &  3   &  14,28,42\\
    \hline
    0.7        &   0.3   &  0.87 s  &  0.24   &  3   &  15,30,45\\
    \hline
    0.5        &   0.5   &  0.81 s  &  1.58  &  2   &  18,36\\
    \hline
    0.3        &   0.7   &  0.84 s  &  2.72 &  2   &  19,38\\
    \hline
    0.1        &   0.9   &  0.81 s   &  11.25  &  2   & 22,44\\
    \hline
    \end{tabular}
    \caption{Effect of weights $W_{d}$ and $W_{s}$ for RL. }
    \label{tb:Effect of weightss_rl}
\end{table}

\subsubsection{Impact of Process Noise}
For the capture column, the inlet flow rate and concentration of mAb from the buffer tank are identified as disturbances, and the noises were simulated with a mean of 0 and a standard deviation ranging from 2 to 4. For each noise level, 5 simulations were conducted, and the average production loss and total cost are reported in Table~\ref{tb:Impact of process noise for RL}. With noisier disturbances, both the production loss and total cost increased, indicating that the performance of the trained RL controller was affected by the presented disturbances. The switching steps are summarized in the last column of the table, and it shows that the periodic pattern in Table \ref{tb:Effect of weightss_rl} is not preserved anymore due to the stochasticity. With a higher noise standard deviation, the first step to switch happened at or after the $15^{th}$ step, resulting in more production loss.
%Song compared with EMPC case, we have a monotonically increase in production loss and total cost whereas EMPC was robust with respect to the process disturbance.

\begin{table}[t]
\normalsize
    \centering
    \begin{tabular}{|c|c|c|c|c|c|}
    \hline
     Noise$_{std} $   &  $\mathcal{PL}_{ave}(mg/L)$  &  $\mathcal{TC}_{ave}$ &  Steps  \\
    \hline
    \hline
    2        & 0.39    &   1.69    &  15,30,45\\
             &         &           &  15,30,47\\ 
             &         &           &  15,31,46\\
             &         &           &  15,31,47\\
            &          &           &  16,31,46\\
    \hline  
    3        & 0.40 &  1.70        &  15,30,44\\
             &         &           &   15,31,47\\
             &         &           &   16,29,44\\
             &          &           &  16,31,48\\
             &          &           &  16,32,47\\     
    \hline
    4       & 0.63     & 1.71        &   15,30,44\\
            &          &             &   15,31,47\\
            &          &             &   16,29,44\\
            &          &             &   16,31,48\\
            &          &             &   18,36\\
    \hline
        5       & 0.71  & 1.75       &   15,30,44\\
            &          &             &   15,32,49\\
            &          &             &   16,29,44\\
            &          &             &   16,31,48\\
            &          &             &   18,36\\
    \hline
    \end{tabular}
    \caption{Effect of process noise for RL.}
    \label{tb:Impact of process noise for RL}
\end{table}

\subsubsection{Impact of Measurement Noise}
The study of the impact of measurement noise on RL performance was conducted, and the experimental setup and results are summarized in Table~\ref{tb:Impact of measurement noise for RL}. Interestingly, given that the magnitude of $\alpha$ is higher than that in the EMPC case, the results for the 3 cases are the same as the deterministic case. This implies that the effect of the measurement noise is negligible with the given $\alpha$ and standard deviation. One possible reason is that the policy obtained by RL takes the whole state space as the input. When only the measured state, $x_{out}$, is subjected to the measurement noise, the modification to the state space is minimal. On the contrary, EMPC, as expressed in Equation~\eqref{eq:Relu optimization}, explicitly considers $x_{out}$ and hence will be more responsive to the measurement noise.
%Song: discuss with Bernard
\begin{table}[t]
\normalsize
    \centering
    \begin{tabular}{|c|c|c|c|c|c|}
    \hline
    $\alpha$    & Noise$_{std} $   &   $\mathcal{PL}_{ave}(mg/L)$  &  $\mathcal{TC}_{ave}$ &  Steps  \\
    \hline
    \hline
    $1$ &   2        & 0.24 & 1.62 &   15,30,45\\
    \hline
    $1$ &   3        & 0.24 & 1.62 &   15,30,45\\
    \hline
    $1$ &   4       & 0.24 & 1.62   &   15,30,45\\
    \hline
    \end{tabular}
    \caption{Effect of measurement noise for RL.}
    \label{tb:Impact of measurement noise for RL}
\end{table}

\section{Conclusions}

The analysis of the simulation results from the different approaches provides valuable insights into the performance and effectiveness of each approach. The proposed approaches outperformed the traditional approach in terms of product loss and cost optimization. 
The sigmoid approach, while saving more product at the column outlet, resulted in higher cost and took a significantly longer time for computation. On the other hand, the ReLU approach showed a balance between product loss and cost, outperforming the RL approach in terms of product loss. However, the RL approach exhibited the best results with the lowest overall cost.

The impact of different factors such as prediction horizon, noise, and the weights on product loss and switching action on the simulation results were investigated on the ReLU, revealing that a smaller horizon led to delayed switching, increased product loss, and higher cost. Increasing the horizon allowed for earlier switching, resulting in reduced product loss and lower cost. Based on these findings, a horizon value of 10 was selected as a suitable compromise between minimizing product loss and cost while maintaining timely decision-making.

The impact of varying the weights, $W_s$ and $W_d$, was thoroughly examined to understand their influence on the simulation results. The analysis revealed that the selection of weights significantly affects the system behavior, particularly in terms of the number of switches and product loss. A smaller $W_d$ allows for more frequent switching, resulting in reduced product loss. Conversely, higher values of $W_s$ and lower values of $W_d$ lead to a smaller number of switches and minimal product loss. Considering these findings, the weights $W_s$ and $W_d$ were carefully chosen as 0.7 and 0.3, respectively. A weight of 0.7 for $W_s$ emphasizes the importance of minimizing product loss. On the other hand, a weight of 0.3 for $W_d$ strikes a balance between minimizing unnecessary switches and optimizing resource utilization. This approach ensures efficient decision-making that aligns with the dual objectives of minimizing product loss and maximizing resource efficiency.

The effect of process noise was also analyzed, indicating that the presence of noise disrupted the periodic switching behavior observed in the noise-free cases, with fluctuations in product loss and cost due to the influence of the noise. Additionally, the impact of measurement noise was examined, showing that the system exhibited robustness in maintaining a periodic switching pattern even under higher levels of measurement noise. The average product loss remained relatively constant, highlighting the controller's ability to adapt to the noise. However, at much higher noise levels for the ReLU case, deviations from the expected behavior were observed, indicating potential challenges in the controller's performance.

For the studies on RL, a similar observation was made while varying $W_s$ and $W_d$ as in the ReLU case. Regarding the impact of process noise, worse economic performance was observed as higher process noise was introduced to the system. However, in the case of measurement noise, since the noise was applied to only one state out of the full state space, the RL agent did not detect it, resulting in the same results as the deterministic case.
By understanding these factors and their impact, more efficient control strategies can be developed to enhance the system's performance and robustness.

In conclusion, the proposed approaches demonstrate improved performance compared to the traditional approach. The analysis of different factors provides valuable insights for optimizing control strategies and addressing challenges such as process and measurement noise. This research contributes to the development of effective control methods in the context of the studied system, with potential applications in various industries.

\section{Acknowledgment}
The authors would like to acknowledge financial support from the Natural Sciences and Engineering Research Council (NSERC) of Canada.

%\bibliographystyle{ieeetr}
%\bibliography{Control_refs}
\end{document}